%% file: paper.tex
\documentclass[]{TEAI}
\usepackage{helvet}

\usepackage{amsmath}
\usepackage{natbib}
\usepackage{graphicx}
\usepackage{subcaption}

\usepackage[toc,page,header]{appendix}
\usepackage[utf8]{inputenc}
\usepackage[T1]{fontenc}
\usepackage{hyperref}
\usepackage{url}
\usepackage{booktabs}
\usepackage{lmodern}
\usepackage{amsfonts}
\usepackage{nicefrac}
\usepackage{microtype}
\usepackage{wrapfig}

\usepackage{amssymb}
\usepackage{fontawesome}
\providecommand{\Envelope}{\faEnvelopeO} % outline envelope marker for corresponding author

\usepackage{titletoc}
\usepackage{tikz}
\usepackage{comment}
\usepackage{tabularx}

\usepackage{minitoc}

\usepackage{array}
\usepackage{etoolbox}

\definecolor{lightblue}{RGB}{200, 230, 255}
\definecolor{headerblue}{RGB}{150, 200, 255}

\usepackage{pgfplots}
\usepackage{xcolor}
\usepackage{float}
\usepackage{multirow}
\usepackage{makecell}
\usepackage{siunitx}
\usepackage{pgf-pie}
\usepackage[export]{adjustbox}

\usepackage{ragged2e}
\usepackage{caption}
\usepackage{enumitem}
\usepackage{pifont}
\usepackage[hang,flushmargin]{footmisc}

\usepackage{tcolorbox}
\tcbuselibrary{breakable}
\tcbuselibrary{skins}

\usepackage{listings}

\usepackage{algorithm}
\usepackage{algpseudocode}

\newcommand{\best}[1]{\textbf{#1}}
\newcommand{\secondbest}[1]{\underline{#1}}

\title{EvoMemNav: Efficient Self-Evolving Fine-Grained Memory for Zero-Shot Embodied Navigation}

\renewcommand{\authorfont}{\fontsize{11}{13}\selectfont}

\author{
    \mbox{Zuhao Ge\textsuperscript{1,2}},
    \mbox{Xiaosong Jia\textsuperscript{1,2,\Envelope}},
    \mbox{Chao Wu\textsuperscript{1,2}},
    \mbox{Yuchen Zhou\textsuperscript{1,2}},
    \mbox{Zuxuan Wu\textsuperscript{1,2}},
    \mbox{Yu-Gang Jiang\textsuperscript{1,2}}
}

\affiliation[1]{\mbox{Institute of Trustworthy Embodied AI (TEAI), Fudan University}}
\affiliation[2]{\mbox{Shanghai Key Laboratory of Multimodal Embodied AI}}

\abstract{
Building memory is essential for long-horizon planning in zero-shot embodied navigation. Detector-centric scene graphs often compress observations into sparse nodes, discarding fine-grained visual evidence and accumulating noise, while 3D reconstruction-based methods remain computationally prohibitive. We present EvoMemNav, an efficient, self-evolving, fine-grained memory framework for zero-shot embodied navigation.
EvoMemNav constructs a Visual-Semantic Memory Graph (VSMGraph) that keeps raw views as first-class memory and organizes them with lightweight semantic cues and topological relations into a room-view-object hierarchy, preserving fine-grained details for disambiguation and \textsc{Stop} verification. To scale to growing memory, we introduce a budgeted coarse-to-fine policy: a coarse stage compresses the search space into promising regions, and a fine stage invokes a VLM only for targeted verification and decision. Beyond static memories, EvoMemNav performs reflection-driven write-back after each subtask, updating graph-attached priors that encode accumulated environmental knowledge to refine future decisions without retraining. Experiments on GOAT-Bench and HM3D across object, text-description, and image-goal modalities show consistent gains in SR/SPL, with better multi-instance disambiguation, fewer premature stops, and stronger zero-shot generalization.
}

\begin{document}
\maketitle
\renewcommand{\thefootnote}{}
\footnotetext{\Envelope~Corresponding author.}
\footnotetext{Code: \href{https://github.com/caicaiya123/EvoMemNav}{\faGithub~github.com/caicaiya123/EvoMemNav}}
\renewcommand{\thefootnote}{\arabic{footnote}}

\vspace{-1.5em}

\input{sec/1_intro}

\input{sec/2_Related}

\input{sec/3_Method}
\input{sec/4_experiments}

\input{sec/5_conclusion}

\clearpage

\bibliographystyle{plainnat}
\bibliography{main}

\end{document}

%% file: sec/1_intro.tex
\section{Introduction}
\label{sec:intro}
In real-world applications, such as home assistants and service robots, mobile agents are expected to understand user instructions, perceive environments, and navigate to target objects~\cite{survey1}. Recent progress in Large Language Models (LLMs) and Vision-Language Models (VLMs) has spurred zero-shot embodied navigation methods~\cite{survey2}.
A key module in these methods is the memory system, which enables long-term reasoning rather than relying solely on short-term, view-by-view responses.

\begin{figure}[t]
  \centering
  \includegraphics[width=\linewidth]{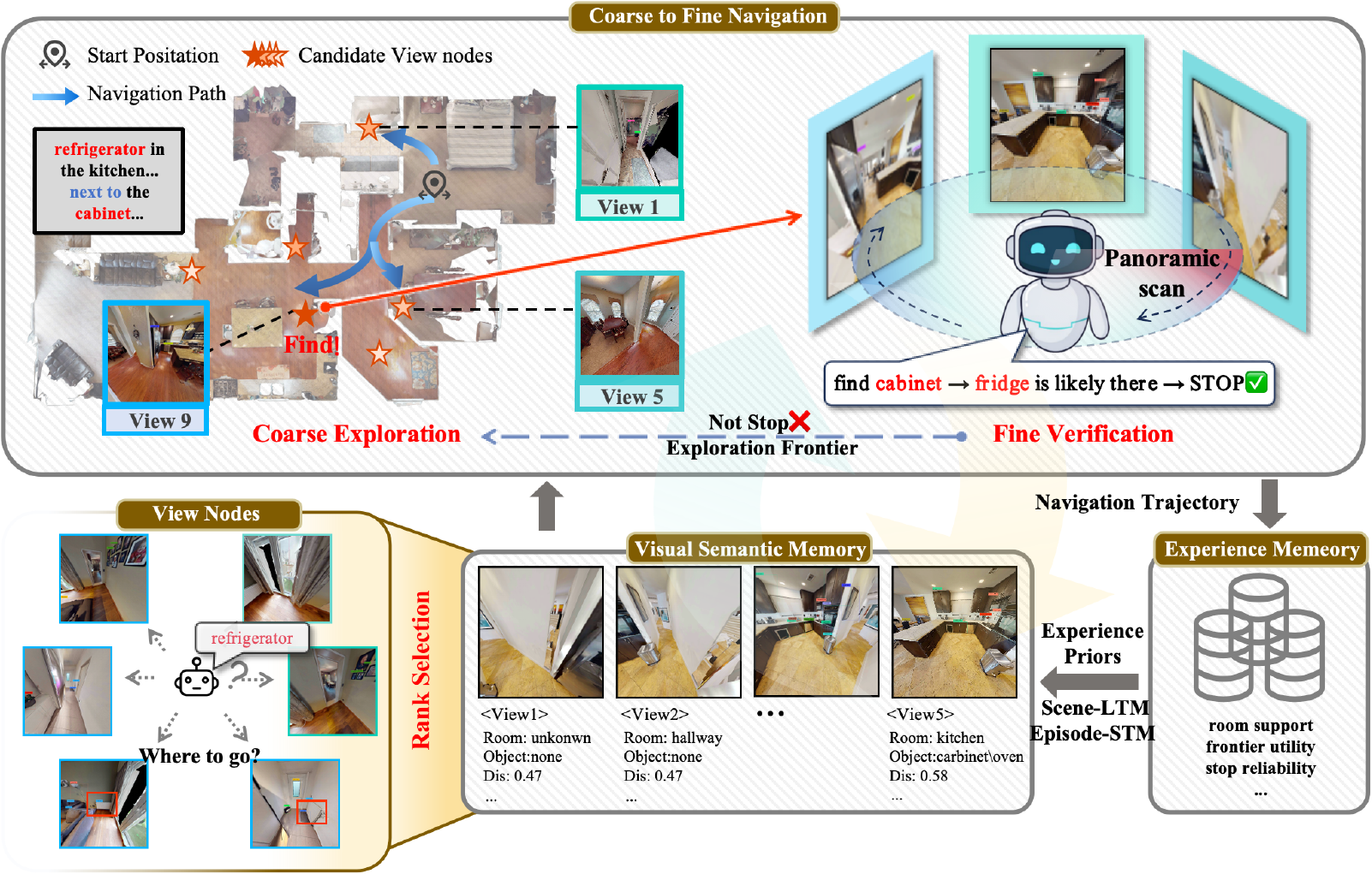}
\caption{\textbf{Overview of EvoMemNav.}
(A) VSMGraph organizes memory into a room-view-object graph with first-class view nodes and lightweight object/room cues for retrieval.
(B) A budgeted coarse-to-fine policy compresses candidates (bucketing + Top-$K$) and invokes a VLM only on the shortlist for multi-view \textsc{Stop} verification.
(C) Reflection-driven write-back updates graph-attached priors (Episode-STM/Scene-LTM) for continual self-evolving improvement.}
   \vspace{-6mm}
  \label{fig:intro}
\end{figure}

% Dector错误影响无法解决，只能vlm缓解（兼顾效率）
To construct memory, a prominent way is to build scene graphs, whose nodes are detected objects/rooms and edges are their geometric relationships, sometimes combined with open-vocabulary retrieval~\cite{clip,clio,cg,roboexp,hovsg}.
However, \textbf{object-centric scene graph memory} (i) compresses observations into sparse symbols and discards fine-grained cues such as textures and spatial layouts; (ii) is sensitive to detection noise that accumulates through downstream reasoning; and (iii) prevents powerful VLMs from directly reasoning over multi-view images. Alternatively, \textbf{3D-reconstruction-based memory}~\cite{survey3} is proposed, yet it incurs high computational overhead at runtime and lacks native compatibility with VLMs.

In contrast, image-based memory retains rich evidence by storing views as nodes and connecting them through topological adjacency~\cite{mapgpt,mobilityVLA,3dMem,msgnav}. Nevertheless, existing methods have two practical limitations: (i) \textbf{Images are loosely organized in a cache}, lacking structured representations for rooms, frontiers, or reachability; consequently, premature termination frequently occurs on same-category objects, and region-level decisions are difficult to articulate. (ii) \textbf{The memory component lacks evolution}: successes and failures are not fed back into reusable states, leading to repeated errors.

% In this work, we present \textbf{EvoTopoNav}, an organized, efficient, and self-evolving image based memory pardigm for zero-shot VLN. The proposed Visual-Semantic Topological Memory Graph (VSTMG) organizes memory images according to both topology and semantics.
% We propose to use anchor views and frontier views as nodes, representing object-clustered evidence and reachable boundaries respectively. 
% This structure enables efficient retrieval of a compact candidate set and supports structured, VLM-in-the-loop reasoning over multi-view images, eliminating the need for constructing heavy full 3D maps.

EvoMemNav builds a \textbf{Visual-Semantic Memory Graph (VSMGraph)} that keeps raw views as first-class memory and organizes them with lightweight semantic cues and topological relations into a structured \emph{room-view-object} hierarchy.
We instantiate two types of view nodes: \emph{anchor views} that store object-rich raw evidence regions for verification
and \emph{frontier views} that represent reachable boundaries for exploration. 
Importantly, object cues (optionally from a lightweight instance cache) are used only as \emph{soft tags} for candidate bucketing and \textsc{Stop} verification; high-level decisions remain grounded in multi-view image evidence on the VSMGraph, avoiding reliance on heavy 3D reconstruction.
In other words, we still maintain object hypotheses as graph nodes for indexing, but we do not compress memory into them:
view nodes preserve raw evidence, and all \textsc{Stop} decisions are verified by multi-view images.

To scale to growing memory and reduce expensive VLM reasoning, we propose a \textbf{budgeted coarse-to-fine navigation policy} operating on the VSMGraph.
The coarse stage (\emph{Explore}) compresses the search space into a small set of promising anchors/frontiers via structured bucketing and ranking, while the fine stage (\emph{Search} + \emph{Verify}) invokes a VLM only for targeted shortlist selection and multi-view \textsc{Stop} verification.
When the decision is uncertain, the policy prioritizes further exploration to avoid hard guessing in multi-instance settings.

Beyond static memories, EvoMemNav performs \textbf{reflection-driven write-back} after each subtask, updating \emph{graph-attached priors} that encode accumulated environmental knowledge (e.g., room support, frontier utility vs.\ dead-end risk, and anchor-view stop reliability).
These priors are attached to room/view/frontier nodes and directly influence future candidate ranking and local verification, enabling continuous improvement without retraining.

Experiments on GOAT-Bench and HM3D across object, description, and image-goal modalities show consistent gains in SR/SPL, with better multi-instance disambiguation, fewer premature stops, and stronger zero-shot generalization.

%% file: sec/2_Related.tex
\begin{figure}[tb]
  \centering
  \includegraphics[width=\linewidth]{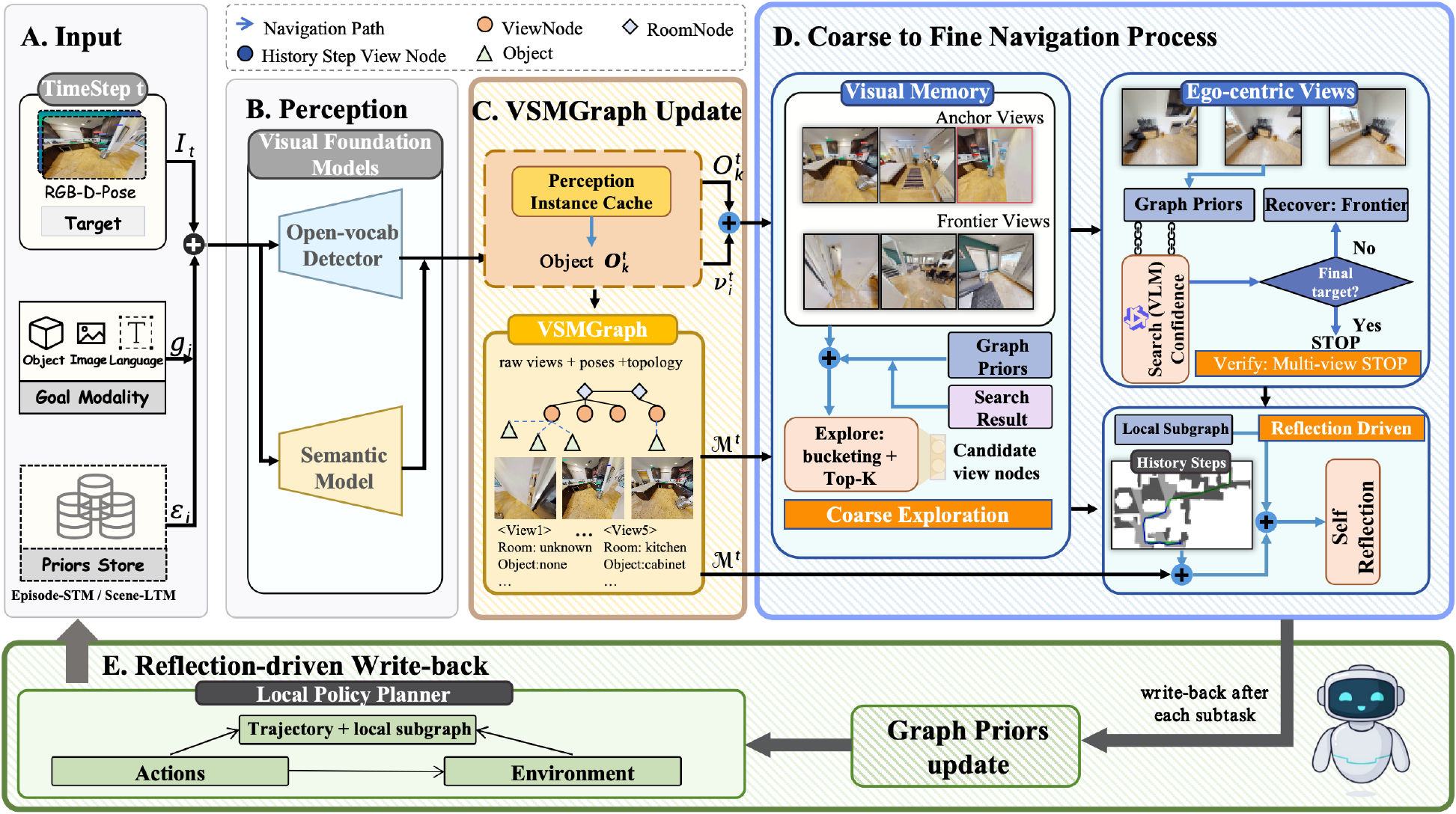}
\caption{\textbf{Framework of EvoMemNav.}
EvoMemNav maintains an image-grounded \textbf{VSMGraph}, a room-view-object memory graph where view nodes store raw image evidence.
A \textbf{budgeted coarse-to-fine} controller shortlists candidates and queries a VLM only on the shortlist for targeted decision and multi-view image \textsc{Stop} verification.
Reflection-driven write-back updates \emph{graph-attached priors} (Episode-STM/Scene-LTM) after each subtask, enabling self-evolving navigation without retraining.}
\label{fig:evoTopoNav}

\end{figure}

\section{Related Work}
\subsection{Visual Navigation}
Classical visual navigation methods are typically categorized as either end-to-end learning~\cite{onEarly,onRL} or modular pipelines~\cite{gose,topoSlam,InstanceImageNavigation,poni,mopa}.
End-to-end approaches require extensive training and often exhibit limited generalization, while modular pipelines reduce the training burden but continue to depend on task-specific policies.
Recent studies have incorporated Large Language Models (LLMs)~\cite{gpt4,llama} and Vision-Language Models (VLMs)~\cite{qwen3,internvl3} to facilitate \emph{zero-shot} navigation~\cite{zson,esc,cows,instructnav,voronav,trihelper,3dMem,emrCOTNav,cogNav,clCOTNav,stairway}.

% Building on these advancements, we adhere to the same objectives for zero-shot navigation and extend them by introducing a Visual-Semantic Topological Memory Graph and an uncertainty-aware coarse-to-fine exploration policy.

\subsection{Scene Representations in Visual Navigation}
Embodied navigation utilizes a variety of representational frameworks.
(1) Metric or 3D maps with semantic information enhance geometric point clouds by incorporating open-vocabulary features~\cite{ok-robot,Ovl-map,beliefmap}.
(2) Detector-driven scene graphs represent objects and rooms as nodes connected by textual relations~\cite{birdeye,cg,H3dsg,DH3dsg,voronav,unigoal,sgnav,emrCOTNav}.
(3) Image-based topological graphs represent visual observations as nodes, which are connected based on spatial adjacency~\cite{mobilityVLA,robohop}.
MapGPT~\cite{mapgpt} employs map-guided prompts, while 3D-Mem~\cite{3dMem} maintains multi-view snapshots and records glimpses of unexplored regions.

\subsection{Memory in Visual Navigation}
% \vspace*{-0.6\baselineskip}
% \noindent 
Zero-shot LLM/VLM navigation agents typically store information using explicit maps.
Empirical evidence from lifelong navigation indicates that the content of memory significantly influences performance on subsequent tasks~\cite{goatbench}.
Attempts to persist memory include storing multi-view snapshots/frontiers~\cite{3dMem,msgnav} or designing long-short memory systems~\cite{apexnav,mem4Nav}.
However, these methods do not incorporate mechanisms to distill trajectories into reusable experiences, resulting in memory accumulation without corresponding improvement.
Reflection mechanisms in LLM agents, ranging from verbal feedback buffers~\cite{evolve1} to strategy-level reviews~\cite{evolve2}, demonstrate potential but remain underexplored in the context of embodied navigation with topological memory.

%% file: sec/3_Method.tex
\section{Methodology}

\subsection{Problem Formulation}
% cvpr版：
% The lifelong embodied navigation task~\cite{goatbench} requires an agent to navigate within an unknown environment to locate an object belonging to a specified category $c$ (e.g., chair, bed). This task is multimodal, meaning that the target is specified using one of three modalities: a category label, an image, or a text description.
% In this task setting, each episode consists of a sequence of subtasks, with the goal of the $k_{th}$ subtask denoted as $g^k$. Each episode begins by placing the agent at a randomized starting pose within an unseen scene $\mathcal{S}$.
% At each time step $t$, the agent receives a posed RGB-D image $I_t = \left\langle I_t^{\mathrm{rgb}}, I_t^{\mathrm{depth}}, p_t \right\rangle$, where the pose $p_t$ comprises both location and orientation, defined as $p_t = (x_t, y_t, \theta_t)$.
% The task is considered successful if the agent stops within 1.0 meters of the target object within a time budget of 500 steps.
% The next subtask goal $g^{k+1}$ is provided only after the current subtask is completed.
% Importantly, the task is open-vocabulary, meaning the agent is not restricted to a fixed set of object categories.

Lifelong embodied navigation~\cite{goatbench} in an unseen scene $\mathcal{S}$ is a task where an agent must solve a sequence of subtasks with multi-modal goals $g^k$ (object label, reference image, or text description). The agent starts from a random pose and receives posed RGB-D observations $I_t = \langle I_t^{\mathrm{rgb}}, I_t^{\mathrm{depth}}, p_t \rangle$, where $p_t = (x_t, y_t, \theta_t)$. A subtask terminates when the agent issues \textsc{stop} or reaches the step budget; it is successful if \textsc{stop} is issued within $1.0\,\mathrm{m}$ of a goal-matching instance.

\subsection{Overview}

As shown in Fig.~\ref{fig:evoTopoNav}, EvoMemNav takes a multi-modal goal $g$ (object label, text, or image) and a stream of posed RGB-D observations $I_t$ as inputs.
It maintains an image-grounded \textbf{Visual-Semantic Memory Graph (VSMGraph)} $\mathcal{G}^t$ with a room-view-object hierarchy:
view nodes store raw observations and poses as first-class \emph{image evidence}, object nodes store lightweight instance hypotheses,
and they are connected by navigability and visibility relations.
View nodes are partitioned into \emph{anchor views} (object-rich evidence regions) and \emph{frontier views} (reachable exploration boundaries).
Each view is additionally annotated with lightweight semantic cues (e.g., room tags and object visibility tags) that serve as retrieval signals;
final decisions are grounded in multi-view image evidence on the VSMGraph rather than detector outputs or dense 3D reconstruction.

On top of $\mathcal{G}^t$, we employ a \textbf{budgeted coarse-to-fine} navigation policy that limits expensive VLM reasoning to a small shortlisted set.
The \textbf{coarse stage} (\emph{Explore}) compresses the candidate space and routes the agent either to a frontier view (exploration) or to an anchor view (evidence), while the \textbf{fine stage} (\emph{Search} + \emph{Verify}) queries the VLM only on this shortlist to select the next view and to perform structured multi-view \textsc{Stop} verification.
A lightweight \emph{Recover} fallback is triggered only when repeated reselect or cooldown indicates a deadlock, temporarily enforcing frontier-only exploration.

Finally, we introduce \textbf{reflection-driven continual memory adaptation (RDCMA)}, a training-free mechanism that writes back subtask outcomes as goal-conditioned, graph-attached priors to enable self-evolving navigation; details are deferred to Sec.~\ref{sec:rdcma}.

\begin{figure}[t]
  \centering
  \includegraphics[width=\linewidth]{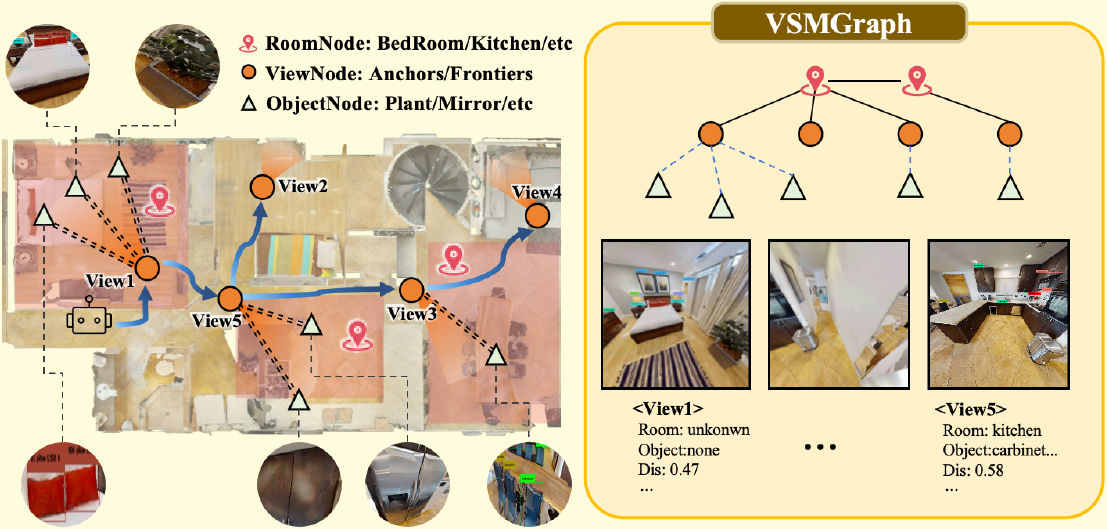}
  \caption{\textbf{VSMGraph construction and attributes.}
Left: as the agent explores on an occupancy grid, we add view nodes along the trajectory and frontier view nodes at reachable boundary regions; object instances are maintained in an object-centric map $\mathcal{O}^{map}$ and linked to views via visibility relations.
Right: the resulting VSMGraph organizes \emph{room-view-object} hierarchy. Each view node stores raw image evidence and pose, and is annotated with a discrete CLIP room-type label used for room-aware bucketing (candidate compression). Solid edges denote navigability and dashed edges denote visibility.}
  \label{fig:graph}
\end{figure}

\subsection{Visual-Semantic Memory Graph Construction}

\subsubsection{VSMGraph Representation}

As shown in Fig.~\ref{fig:graph}, we maintain a Visual-Semantic Memory Graph (VSMGraph)
$\mathcal{G}^t=(\mathcal{R}^t,\mathcal{V}^t,\mathcal{O}^t,\mathcal{E}^t)$ with a room-view-object hierarchy.

Room nodes $\mathcal{R}^t$ index views and objects within the same region, enabling region-level retrieval.
View nodes $\mathcal{V}^t$ store \emph{image evidence} and poses, while object nodes $\mathcal{O}^t$ store \emph{instance hypotheses} (category and 3D localization).
Edges $\mathcal{E}^t$ include (i) \emph{navigability} edges between successive collision-free viewpoints and (ii) \emph{visibility} edges linking a view to the object nodes observed from it.

We update $\mathcal{G}^t$ online. At each visited pose, we add a view node and connect it to the previous view node by a navigability edge.
An occupancy grid serves as the metric scaffold: shortest paths are computed on the grid and executed as walks along view-to-view edges, while high-level reasoning and retrieval operate on $\mathcal{G}^t$.
To support efficient candidate compression in our budgeted coarse-to-fine policy (Sec.~\ref{sec:coarse_to_fine}), each view is annotated with a discrete room-type label
$\rho(v)\in\mathcal{T}_{room}$ computed by CLIP $\arg\max$ matching between the view image and a set of common room types, which serves as a lightweight, threshold-free contextual tag for room-aware bucketing.

Besides, VSMGraph exposes room/view/frontier identifiers and attributes (e.g., $\rho(v)$ and visibility-derived object-hit tags), and RDCMA attaches lightweight priors to these elements (Sec.~\ref{sec:rdcma}).

\subsubsection{Anchor Views}
We partition view nodes into anchor and frontier views, i.e.,
$\mathcal{V}^t=\mathcal{V}^{A,t}\cup \mathcal{V}^{F,t}$.
An anchor view captures an object-rich explored observation:
\[
v_t^{A}=\langle I_t^{\mathrm{rgb}},\, p_t,\, \mathcal{O}_t^{\mathrm{vis}},\, \rho_t^{\mathrm{room}}\rangle,
\]
where $\rho_t^{\mathrm{room}}$ is the CLIP room label and $\mathcal{O}_t^{\mathrm{vis}}$ is the set of visible object hypotheses/tags associated with this view.

In practice, we maintain an object-centric cache $\mathcal{O}^{map}$ of detected instances
to derive view-level visibility tags $\mathcal{O}_t^{\mathrm{vis}}$.
\emph{This cache does not define the VSMGraph: it only annotates view nodes with lightweight cues for retrieval, while final decisions are verified by multi-view image evidence.}
Suppressing these tags therefore degrades candidate ordering and exploration efficiency rather than breaking the closed-loop controller, since \textsc{Stop} verification remains grounded in raw multi-view images.

\subsubsection{Frontier Views}
We adopt frontier-based exploration on a top-down occupancy grid $\mathbf{M}_t$.
A frontier cell is a free cell adjacent to at least one unknown cell; frontier cells are clustered into frontier regions.
For each region $f$, we choose a reachable entry pose $p_f$ and capture a frontier-facing image $I_f^{\mathrm{front}}$, defining a frontier view as
\[
F_f=\langle r_f,\, p_f,\, I_f^{\mathrm{front}},\, \rho_f^{\mathrm{room}}\rangle,
\]
where $r_f$ denotes the frontier region on the grid and $\rho_f^{\mathrm{room}}$ is the CLIP room label.
Each frontier view is connected to the closest explored view (and its room node), so unexplored areas appear as boundary nodes at the periphery of $\mathcal{G}^t$.

\subsection{Coarse-to-Fine Navigation Policy}
\label{sec:coarse_to_fine}
We cast navigation as traversal on the VSMGraph $\mathcal{G}^t$: at time $t$, the agent at view node $v_t$ selects the next target node to navigate to.
A common zero-shot pipeline filters anchor views by target cues, queries a VLM over a large and growing candidate pool, and often stops immediately once an anchor view is selected. This design (i) scales poorly with memory size, (ii) provides limited structure for comparing exploration frontiers, and (iii) is prone to same-class wrong-instance and premature-stop failures.
To address these issues, we adopt a \emph{budgeted coarse-to-fine} framework that separates \textbf{candidate compression} from \textbf{fine-grained VLM reasoning}, so that expensive VLM calls are reserved for a small set of candidates and for \textsc{Stop} verification.

\paragraph{Budgeted coarse-to-fine.}
We use \emph{coarse-to-fine} to denote a budgeted two-stage decision process:
the \textbf{coarse stage} (\emph{Explore}) compresses the candidate space and routes the agent to either a \emph{frontier view} (exploration) or an \emph{anchor view} (evidence), while
the \textbf{fine stage} (\emph{Search} + \emph{Verify}) performs small-set selection and structured \textsc{Stop} verification.

\subsubsection{Coarse stage (Explore): candidate compression and routing}
Given goal $g$ and graph $\mathcal{G}^t$, we construct compact candidate sets
$\mathcal{C}_t^A \subset \mathcal{V}^{A,t}$ (anchor views) and $\mathcal{C}_t^F \subset \mathcal{V}^{F,t}$ (frontiers),
with budgets $|\mathcal{C}_t^A|\le K_A$ and $|\mathcal{C}_t^F|\le K_F$.
Candidates are compressed via room-aware bucketing under soft constraints, using the discrete room label $\rho(\cdot)$ and lightweight object-hit tags derived from $\mathcal{O}^{vis}$.
Optionally, a single \emph{text-only} query to the same VLM provides coarse focus hints that are used only as soft cues during bucketing.
If the compressed anchor set is empty ($\mathcal{C}_t^A=\varnothing$), we directly route to a frontier without querying the VLM.

\subsubsection{Fine stage (Search + Verify): small-set selection and \textsc{Stop} verification}
% We prompt the VLM with the goal $g$ and the \emph{raw images} of shortlisted anchor/frontier views (plus minimal tags such as room labels for context).
In the \emph{Search} step, we query the VLM only over the compact set $\mathcal{C}_t=\mathcal{C}_t^A\cup\mathcal{C}_t^F$:
\begin{equation}
(a_t, \sigma_t) = \mathrm{VLM}(g,\, \mathcal{C}_t), \qquad \sigma_t \in \{\texttt{certain}, \texttt{uncertain}, \texttt{unknown}\}.
\end{equation}
If an anchor view is selected but the decision is low-confidence ($\sigma_t\in\{\texttt{uncertain},\texttt{unknown}\}$), we gate back to exploration and route to a frontier instead, reducing same-class wrong-instance errors.
In the \emph{Verify} step, upon reaching a selected anchor view, we do not stop immediately; instead, we invoke structured \textsc{Stop} verification that returns \texttt{STOP} or \texttt{RESELECT}.
A \texttt{RESELECT} verdict triggers anchor-view cooldown and re-enters the coarse stage, forming a robust closed loop for handling premature-stop and wrong-instance cases.

\subsection{Reflection-Driven Continual Memory Adaptation}
\label{sec:rdcma}

% TODO 修改这个图
\begin{figure}[t]
  \centering
  \includegraphics[width=\linewidth]{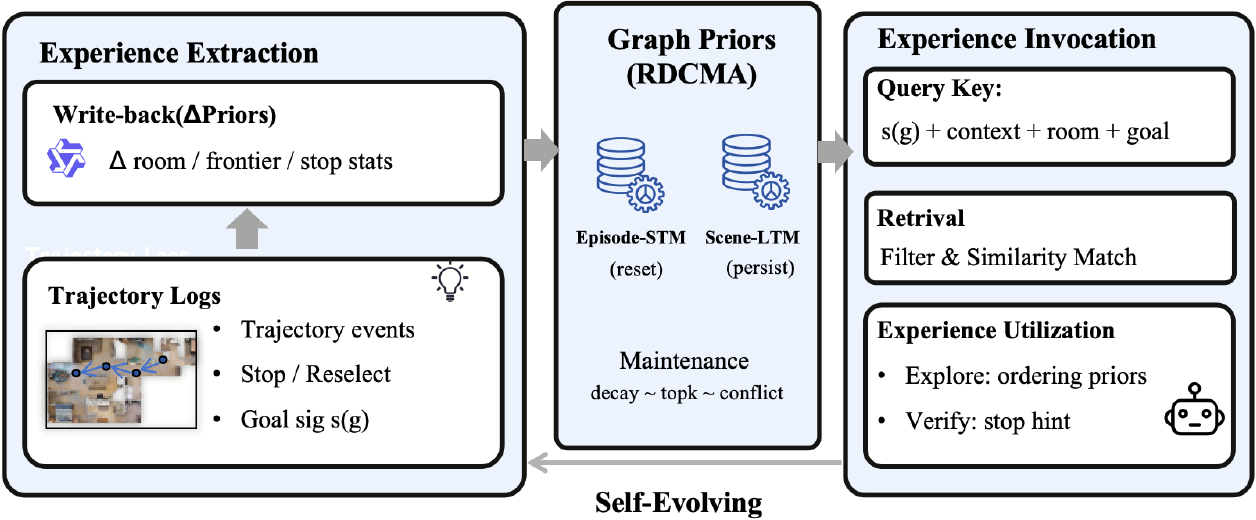}
\caption{\textbf{Reflection-Driven Continual Memory Adaptation (RDCMA).}
A training-free, graph-attached prior mechanism for the VSMGraph: trajectory events and \textsc{Stop} outcomes (\texttt{STOP}/\texttt{RESELECT}) update compact priors under $s(g)$, stored as Episode-STM (reset) and Scene-LTM (persistent) with decay/top-$K$/conflict.
Goal-conditioned lookup applies priors as \emph{Explore} tie-breakers and a \emph{Verify} stop hint, with no parameter updates.}
  \label{fig:self_evolve}
\end{figure}

The lifelong multi-subtask setting enables an agent to reuse evidence accumulated earlier in the same scene.
Beyond the online-growing VSMGraph structure, we introduce \emph{Reflection-Driven Continual Memory Adaptation (RDCMA)}, a training-free mechanism that writes back interaction outcomes as \emph{goal-conditioned priors} attached to VSMGraph elements.
RDCMA maintains priors at two time scales: \textbf{Episode-STM} captures short-term suppressions and loop-avoidance cues within the current episode, while \textbf{Scene-LTM} persists per scene to support continual reuse across subtasks/episodes.
Both memories are indexed by a goal signature $s(g)$ (e.g., modality and/or target class), and support a coarse-to-fine fallback to avoid overly sparse keys.

Priors are \emph{graph-attached}: they are associated with room nodes, anchor views, and frontier views, and stored as compact statistics (e.g., support counts and reliability/risk estimates).
Importantly, RDCMA does not update any model parameters; instead, it continuously updates these lightweight statistics, turning the memory graph into a gradually adapting decision substrate that interfaces cleanly with the coarse-to-fine policy.

\subsubsection{Experience memory via reflection}
RDCMA derives supervision signals directly from the navigation loop.
Concretely, reflection is invoked during \textsc{Stop} verification (to decide \texttt{STOP} vs.\ \texttt{RESELECT}) and the resulting outcomes are aggregated and written back after each subtask.
After each subtask, we summarize interaction outcomes using (i) trajectory events on the VSMGraph (e.g., visited rooms, selected frontiers/anchor views, revisits, and dead-end/loop indicators) and (ii) structured \textsc{Stop}-verification outcomes (\texttt{STOP} vs.~\texttt{RESELECT}) produced by the fine stage.
These signals are then written back as updates to the attached priors under the current goal signature $s(g)$.
Episode-STM is updated online and reset at episode end; Scene-LTM is updated conservatively and persisted per scene.

\subsubsection{Experience retrieval and usage}
RDCMA influences decision-making through two coupled channels.

\paragraph{Exploration Ordering Priors}
During \emph{Explore} (Sec.~\ref{sec:coarse_to_fine}), priors act as conservative tie-breakers within the bucketed candidate set.
They bias frontier selection toward regions with higher historical utility and lower dead-end risk, and bias anchor views selection toward views that have provided reliable evidence under similar goals.
These priors do not override coarse bucketing constraints; they only refine ordering within the bounded candidate set, which improves stability under noisy online feedback.

\paragraph{\textsc{Stop} Calibration Priors}
During \emph{Verify} (Sec.~\ref{sec:coarse_to_fine}), priors calibrate structured \textsc{Stop} decisions.
For each candidate anchor view, we maintain lightweight \textsc{Stop} reliability statistics under $s(g)$ and inject a compact \emph{stop-memory hint} into the verification prompt.
This channel directly targets premature-stop and wrong-instance failures by providing prior evidence about whether an anchor view has been a reliable stopping region for similar goals.

\subsubsection{Maintenance: decay, conflict suppression, and persistence}
To bound memory growth and mitigate contradictory feedback, RDCMA applies simple maintenance: exponential decay for stale priors, a small per-signature cap (retain top-$K$ most useful entries), and conservative suppression of repeatedly contradicted priors. These operations only affect the attached statistics and do not introduce new decision thresholds or override the coarse-to-fine policy.

%% file: sec/4_experiments.tex
\section{Experiments}

\subsection{Experimental Setup}

\subsubsection{Datasets}

%  四个数据集：主要的：Goat Bench，验证泛化性 Hm3d Object Navigation、IIN、TN[后两个再看，似乎做的人不多] -> hm3d 的两个数据集 v1 和 v2 
We evaluate methods in Habitat simulator\cite{habitatsimulator} on large-scale datasets: 

1) GOAT-Bench\cite{goatbench} is a multimodal lifelong navigation benchmark where an agent navigates to objects in unknown scenes, described by a category name, language description, or image. It comprises 36 scenes, each with 10 episodes, containing 5-10 subtasks. Target modalities are uniformly sampled, with targets across all categories on a single floor per episode, in an open-vocabulary setting.

2) The HM3D dataset is a large-scale 3D indoor benchmark for embodied navigation. We follow the standard setup on HM3Dv1~\cite{habitat} (20 scenes, 2000 episodes) and HM3Dv2~\cite{habitat2} (36 scenes, 1000 episodes), both with 6 object-goal categories.

\subsubsection{Metrics}
We report Success Rate (SR) and Success weighted by inverse Path Length (SPL) for all methods, following the mainstream settings.
A navigation task is deemed successful if the agent stops within 1\,m of the navigation goal.
SPL is the success score weighted by exploration distances.
Higher SR and SPL indicate superior performance.

\subsubsection{Implementation Details}
We use Habitat-Lab~\cite{habitatsimulator} for simulation and implement all components in PyTorch.
YOLOv8-World and SAM provide instance hypotheses and visibility-derived tags, which are used as soft cues for candidate bucketing and \textsc{Stop} verification; the core memory remains image-grounded view evidence.
We use Qwen3-VL-8B~\cite{qwen3} as the VLM in both the selection and verification stages. For efficiency, ablations and runtime profiling use a fixed subset covering all scenes (\textbf{episode index=1} for each of the 36 scenes); main results follow the full GOAT-Bench protocol.

\subsection{Quantitative Experiments}
% GoatBench split对比表
% 整体whole set 对比表
% ablation study
% 这里我们在Goat 这个multimodal longlife Embodied Navigation 数据集上做了详细的对比实验。
% goat bench是一个新提出来的非常具有挑战性multimodal longlife 导航数据集，的为了和目前主流的方法，做公平的比较，这里我们也采用了相同的设置，分别报告了在val unseen whole set上的表现，可以发现，相对于baseline而言，我们的方法有着显著的提升。特别是，这里我们并没有采用额外的训练，也没有使用闭源的多模态大模型，整个实验均基于开源的多模态大模型完成。
% 针对goat benchmark数据集的特性，我们还在 VAL UNSEEN Split 数据集上做了验证，这是val unseen 的一个子集，大约在1/10的规模，同样覆盖了36个各不相同的场景，这里我们可以看到我们的方法也取得了非常好的结果，特别是针对图像这个模态而言，可以有效的提升vlm对图像导航的成功率。特别值得注意的是，在三种不同模态的任务重，语言模态往往的结果会比较低，这也意味着语言和图像的对齐，对于多模态大模型而言本身就是一个挑战。

% ---- GOAT-Bench: VAL UNSEEN  whole set----
\begin{table}[tb]
\centering
\setlength{\tabcolsep}{8pt}
\renewcommand{\arraystretch}{1.2}
\caption{\textbf{Comparison on the ``Val Unseen'' split of GOAT-Bench.} ``$\dagger$'' denotes the results we reproduced. \best{Best} in bold, \secondbest{second best} underlined.}
\begin{tabular}{lccc}
\toprule
\textbf{Method} & \textbf{Training-free} & \textbf{SR}\,$\uparrow$ & \textbf{SPL}\,$\uparrow$ \\
\midrule
SenseAct-NN Monolithic~\cite{goatbench}  & $\times$     & 12.3 & 6.8 \\
Modular CLIP on Wheels~\cite{goatbench}  & $\checkmark$ & 16.1 & 10.4 \\
Modular GOAT~\cite{goatbench}            & $\checkmark$ & 24.9 & 17.2 \\
SenseAct-NN Skill Chain~\cite{goatbench} & $\times$     & 29.5 & 11.3 \\
VLMnav~\cite{vlmnav}                     & $\checkmark$ & 20.1 & 9.6 \\
DyNaVLM~\cite{dynavlm}                   & $\checkmark$ & 25.5 & 10.2 \\
TANGO~\cite{tango}                       & $\checkmark$ & 32.1 & 16.5 \\
3D-Mem$^\dagger$~\cite{3dMem}            & $\checkmark$ & 42.6 & 22.8 \\
MTU3D~\cite{mtu3d}                       & $\times$     & 47.2 & 27.7 \\
MSGNav~\cite{msgnav}                     & $\checkmark$ & \secondbest{52.0} & \secondbest{29.6} \\
\midrule
\rowcolor{seedblue!10}
\textbf{EvoMemNav} (Ours)                & $\checkmark$ & \best{59.6} & \best{38.9} \\
\bottomrule
\end{tabular}
\label{tab:goat_bench_results}
\end{table}

% ---- HM3D-Bench: v0.1 & v0.2----
\begin{table}[tb]
\centering
\setlength{\tabcolsep}{8pt}
\renewcommand{\arraystretch}{1.2}
\caption{\textbf{Comparison on the HM3D object-goal navigation benchmark.} \best{Best} in bold, \secondbest{second best} underlined.}
\begin{tabular}{l c cc cc}
\toprule
\multirow{2}{*}{\textbf{Method}} & \multirow{2}{*}{\textbf{Training-free}} &
\multicolumn{2}{c}{\textbf{HM3Dv1}} &
\multicolumn{2}{c}{\textbf{HM3Dv2}} \\
\cmidrule(lr){3-4}\cmidrule(lr){5-6}
& & \textbf{SR}\,$\uparrow$ & \textbf{SPL}\,$\uparrow$
  & \textbf{SR}\,$\uparrow$ & \textbf{SPL}\,$\uparrow$ \\
\midrule
ZSON~\cite{zson}              & $\times$     & 25.5 & 12.6 & --   & --   \\
PixNav~\cite{pixnav}          & $\times$     & 37.9 & 20.5 & --   & --   \\
SPNet~\cite{spnet}            & $\times$     & 31.2 & 10.1 & --   & --   \\
VLFM~\cite{vlfm}              & $\times$     & 52.5 & 30.4 & 63.6 & 32.5 \\
SGM~\cite{sgm}                & $\times$     & \best{60.2} & 30.8 & --   & --   \\
\midrule
ESC~\cite{esc}                & $\checkmark$ & 39.2 & 22.3 & --   & --   \\
L3MVN~\cite{l3mvn}            & $\checkmark$ & 50.4 & 23.1 & 36.3 & 15.7 \\
InstructNav~\cite{instructnav}& $\checkmark$ & --   & --   & 58.0 & 20.9 \\
VLFM*~\cite{vlfm}             & $\checkmark$ & 50.9 & 23.6 & 56.9 & 27.5 \\
GAMap~\cite{gamap}            & $\checkmark$ & 53.1 & 26.0 & --   & --   \\
SG-Nav~\cite{sgnav}           & $\checkmark$ & 54.0 & 24.9 & 49.6 & 25.5 \\
UniGoal~\cite{unigoal}        & $\checkmark$ & 54.5 & 25.1 & --   & --   \\
WMNav~\cite{wmnav}            & $\checkmark$ & 58.1 & \secondbest{31.2} & --   & --   \\
TriHelper~\cite{trihelper}    & $\checkmark$ & 56.5 & 25.3 & --   & --   \\
\midrule
\rowcolor{seedblue!10}
\textbf{EvoMemNav} (Ours)     & $\checkmark$ & \secondbest{59.2} & \best{33.6} & \best{63.8} & \best{39.4} \\
\bottomrule
\end{tabular}
\label{tab:benchmark_Hm3d}
\end{table}

\paragraph{GOAT-Bench.}
Tab.~\ref{tab:goat_bench_results} reports results on GOAT-Bench VAL-UNSEEN.
EvoMemNav achieves \textbf{59.6} SR and \textbf{38.9} SPL, outperforming prior training-free baselines and demonstrating strong generalization across the multi-modal lifelong setting.
Notably, our gains are obtained without task-specific training and with an open-source VLM backbone, indicating that the improvements come from the proposed memory organization and budgeted decision policy.

\paragraph{HM3D ObjectGoal.}
To evaluate generalization beyond GOAT-Bench, we further test on HM3D object-goal navigation (Tab.~\ref{tab:benchmark_Hm3d}).
EvoMemNav achieves \textbf{59.2/33.6} (SR/SPL) on HM3Dv1 and \textbf{63.8/39.4} on HM3Dv2, setting a strong training-free performance and confirming that our method transfers to large-scale object-goal benchmarks.

% 除了定量结果外，我们还提供了案例研究来说明我们的方法的有效性
% 这里，图vis1 展示了VS-TopoMap是如何基于coarse to fine policy 帮助agent 一步一步完成导航任务的，当识别到了错误的目标，fine 推理策略，会利用vlm的推理能力，判断是否应该停下，并对目标物体进行进一步的探索。图vis2 则突出了SE-CFR的稳健性在不同模态目标导航上，当基线失败时，仍然能完成任务，特别是面对同类不同实例的导航，可以高效的在不同场景中探索。

\subsection{Ablation Analysis}
%  失败case 分析
%  给一个横着的比例图（纵轴：baseline、coarsetofine/experience; 横轴：成功；错停；超时；没停准；其他）-> 直观看到每个module的加入，如何拯救失败case的 

% 不同模型基准对比 -> 是否是更好的模型可以解决这些问题？【实验on the way】

% "从粗到细"效率思考？ -> 定义关键帧：是否此时需要查询VLM接口，效率评估标准：调用vlm次数 object、text、Image 【柱状图】

% 经验的可视化，举个case 图，对比【原始的失败；】？【具体如何设计？】
% ---- Ablation on goatbench episode 1----
\begin{table}[tb]
\centering
\setlength{\tabcolsep}{4pt}
\renewcommand{\arraystretch}{1.2}
\caption{\textbf{Ablation on GOAT-Bench VAL-UNSEEN} (episode index=1 per scene; 36 scenes).
All rows share the same perception modules and VLM backbone.
Disabling \textbf{Coarse} falls back to a naive top-$K$ heuristic under the same budgets; disabling \textbf{Fine} causes the agent to stop immediately upon reaching the selected view. \best{Best} overall in bold.}
\begin{tabular}{cccc cc cc cc cc}
\toprule
\multicolumn{4}{c}{\textbf{Settings}} &
\multicolumn{2}{c}{\textbf{Object}} &
\multicolumn{2}{c}{\textbf{Language}} &
\multicolumn{2}{c}{\textbf{Image}} &
\multicolumn{2}{c}{\textbf{Overall}} \\
\cmidrule(lr){1-4}\cmidrule(lr){5-6}\cmidrule(lr){7-8}\cmidrule(lr){9-10}\cmidrule(lr){11-12}
\textbf{VSMGraph} & \textbf{Coarse} & \textbf{Fine} & \textbf{RDCMA} &
\textbf{SR} & \textbf{SPL} &
\textbf{SR} & \textbf{SPL} &
\textbf{SR} & \textbf{SPL} &
\textbf{SR} & \textbf{SPL} \\
\midrule
--          & --          & --          & --          & 49.5 & 27.5 & 34.1 & 18.4 & 35.2 & 19.5 & 39.9 & 22.0 \\
$\checkmark$ & --          & --          & --          & 53.5 & 31.3 & 42.9 & 26.7 & 44.3 & 32.2 & 47.1 & 30.1 \\
$\checkmark$ & $\checkmark$ & --          & --          & 70.7 & 46.0 & \secondbest{54.2} & \best{34.3} & 49.7 & 38.3 & 58.7 & 39.7 \\
$\checkmark$ & --          & $\checkmark$ & --          & 66.7 & 34.2 & 53.9 & 31.8 & 50.0 & 38.4 & 57.2 & 34.7 \\
$\checkmark$ & $\checkmark$ & $\checkmark$ & --          & \secondbest{72.7} & \secondbest{48.3} & 48.4 & 30.4 & \secondbest{60.2} & \secondbest{43.3} & \secondbest{60.8} & \secondbest{40.8} \\
\midrule
\rowcolor{seedblue!10}
$\checkmark$ & $\checkmark$ & $\checkmark$ & $\checkmark$ & \best{75.8} & \best{49.5} & 53.9 & \secondbest{33.2} & \best{63.6} & \best{46.2} & \best{64.8} & \best{43.1} \\
\bottomrule
\end{tabular}
\label{tab:ablation_new}
\end{table}

% \textbf{AppenG} replaces the views cache with the proposed VSTMGraph memory.
% \textbf{Coarse} denotes Explore-stage candidate compression and routing.
% \textbf{Fine} denotes small-set VLM selection plus structured STOP verification.
% \textbf{RDCMA} enables graph attached priors for exploration ordering and STOP calibration.

\subsubsection{Effect of Each Component}

Table~\ref{tab:ablation_new} details the ablation study on four components (\textbf{VSMGraph}, \textbf{Coarse}, \textbf{Fine}, \textbf{RDCMA}) with a fixed VLM backbone. To ensure fair comparison, disabling \textbf{Coarse} falls back to a top-$K$ heuristic under the same budgets, and disabling \textbf{Fine} causes the agent to stop immediately upon reaching the selected view.

Starting from a baseline of 39.9/22.0 (SR/SPL), integrating \textbf{VSMGraph} yields an initial gain (47.1/30.1), validating the benefit of structured visual-semantic memory. Enabling \textbf{Coarse} (\emph{Explore}) provides the most substantial improvement (58.7/39.7), identifying budgeted candidate compression as the primary performance driver. Adding \textbf{Fine} (\emph{Search} + \emph{Verify}) further refines this to 60.8/40.8, notably improving image goals by mitigating premature stops. Finally, \textbf{RDCMA} achieves peak performance (64.8/43.1 Overall SR/SPL), with consistent gains on Object (75.8 SR) and Image (63.6 SR) tasks, demonstrating effective training-free adaptation that complements the coarse-to-fine policy.
We further verify robustness against candidate budgets $(K_A, K_F)$ and VLM backbones in the Supplementary Material, confirming stable performance across configurations.

\subsubsection{Runtime and Budget Analysis}
\label{sec:eff_breakdown}
% ---- Efficient  on goatbench episode 1----
\begin{table}[tb]
\centering
\setlength{\tabcolsep}{4pt}
\renewcommand{\arraystretch}{1.2}
\caption{\textbf{Performance--efficiency on GOAT-Bench VAL-UNSEEN} (episode index=1 per scene; 36 scenes).
All metrics report mean (std) \emph{per subtask}. \textbf{Graph} includes scene-graph memory maintenance. \textbf{Total}\,$\approx$\,VLM\,+\,Graph time. \best{Best} per column in bold.}
\begin{tabular}{l c cc c ccc}
\toprule
\multirow{2}{*}{\textbf{Method}} &
\multirow{2}{*}{\textbf{SR}\,$\uparrow$} &
\multicolumn{2}{c}{\textbf{VLM}} &
\multirow{2}{*}{\textbf{Steps}} &
\multicolumn{3}{c}{\textbf{Time (s)}\,$\downarrow$} \\
\cmidrule(lr){3-4} \cmidrule(lr){6-8}
& & Calls\,$\downarrow$ & Tokens\,$\downarrow$ & & VLM & Graph & Total \\
\midrule
3D-Mem~\cite{3dMem}      & 49.6          & 10.7\,(15.3)         & \best{14.5k\,(18.3k)} & \best{7.0\,(7.7)} & \best{14.0\,(18.5)} & 88.2\,(81.1) & 102.2 \\
VSMGraph+C2F             & 60.8          & \best{6.3\,(1.9)}    & 18.9k\,(6.3k)         & 12.4\,(6.4)       & 25.5\,(7.4)         & \best{37.3\,(21.6)} & 62.8 \\
\rowcolor{seedblue!10}
\textbf{EvoMemNav} (Ours)& \best{64.8}   & 6.5\,(2.3)           & 19.4k\,(7.1k)         & 13.3\,(7.6)       & 21.1\,(6.3)         & 37.6\,(19.9) & \best{58.7} \\
\bottomrule
\end{tabular}
\label{tab:efficiency_main}
\end{table}

% To address concerns that multi-stage VLM reasoning may increase runtime cost, we profile per-subtask efficiency on GOAT-Bench VAL-UNSEEN (36 scenes; episode\_index=1 per scene) and report mean (std) in Tab.~\ref{tab:efficiency_main}. We decompose cost into \textbf{VLM} (calls/tokens and summed wall-clock latency) and \textbf{graph maintenance}; their sum serves as a proxy runtime per subtask.

% \textbf{VSMG+C2F} reduces VLM calls by \textbf{41\%} (10.7$\rightarrow$6.3) and graph time by \textbf{58\%} (88.2$\rightarrow$37.3\,s), cutting total proxy runtime by \textbf{39\%} (102.2$\rightarrow$62.8\,s) while improving SR by \textbf{+11.2} (49.6$\rightarrow$60.8) over 3D-Mem. \textbf{EvoMemNav} further improves SR to \textbf{64.8} and lowers total proxy runtime to \textbf{58.7\,s} with similar graph overhead, indicating that RDCMA adds gains with marginal cost. Although tokens increase due to structured verification prompts, overall efficiency is dominated by fewer VLM calls and lighter graph maintenance. Steps increase as the policy avoids premature stops via verification/reselection; minor components are profiled in the appendix.

To evaluate efficiency, we profile per-subtask costs on GOAT-Bench VAL-UNSEEN (36 scenes), decomposing runtime into \textbf{VLM latency} (calls/tokens) and \textbf{graph maintenance} (Tab.~\ref{tab:efficiency_main}).

Compared to 3D-Mem, \textbf{VSMGraph+C2F} reduces VLM calls by \textbf{41\%} (10.7$\rightarrow$6.3) and graph maintenance by \textbf{58\%} (88.2$\rightarrow$37.3\,s), cutting total proxy runtime by \textbf{39\%} (102.2$\rightarrow$62.8\,s) while improving SR by \textbf{+11.2} (49.6$\rightarrow$60.8). \textbf{EvoMemNav} further boosts SR to \textbf{64.8} and lowers runtime to \textbf{58.7\,s} with marginal overhead, demonstrating that RDCMA yields gains at negligible cost. Although structured verification slightly increases token usage, overall efficiency improves due to fewer VLM calls and lighter graph operations. The moderate step increase reflects the policy's verification-driven reselection, which effectively avoids premature termination (see Supplementary Material for detailed profiling).

\subsubsection{Effect of RDCMA across Subtasks}

\begin{figure}[t]
  \centering
  \includegraphics[width=\linewidth]{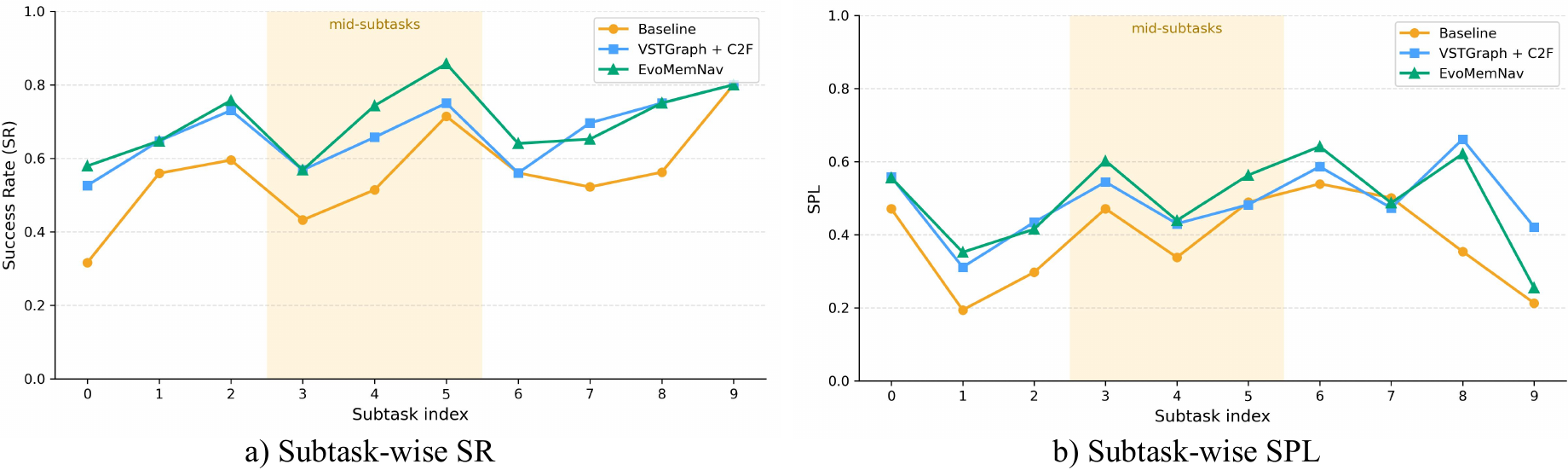}
  \caption{\textbf{Impact of reflection-driven experience memory.} Subtask-wise SR/SPL on GOAT-Bench VAL-UNSEEN. The shaded region (subtasks 3--5) highlights where in-episode experience accumulation yields maximal gains.}
  \label{fig:experience_curve}
\end{figure}

GOAT-Bench's lifelong protocol enables in-scene feedback propagation across subtasks. We evaluate continual adaptation by comparing: (i) baseline, (ii) \textbf{VSMGraph+C2F} (w/o RDCMA), and (iii) \textbf{EvoMemNav} (w/ RDCMA). As shown in Fig.~\ref{fig:experience_curve}, \textbf{VSMGraph+C2F} consistently outperforms the baseline, validating the topology-aware budgeted policy. 

Enabling \textbf{RDCMA} yields progressive gains that correlate with accumulated experience, peaking on mid subtasks (3--5) where \textsc{Stop}-verification and exploration feedback have sufficiently updated graph-attached priors. Notably, improvements on early subtasks indicate effective reuse of scene-level priors when available. Overall, RDCMA provides complementary, training-free self-improvement that enhances VSMGraph+C2F without additional optimization.

\subsection{Qualitative Study}
\begin{figure}[tb]
  \centering
  \includegraphics[width=0.7\linewidth]{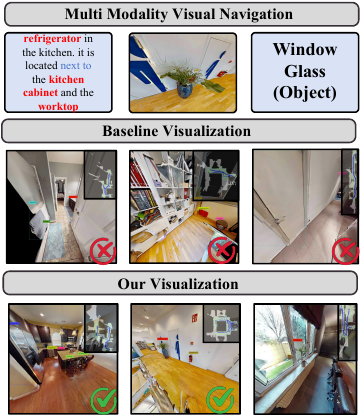}
  \caption{\textbf{Qualitative comparison} with the baseline on GOAT-Bench. From left to right: text, image, and object goal subtasks. EvoMemNav retrieves compact candidates from the VSMGraph, plans on the top-down map, and verifies \textsc{Stop} decisions, reducing wrong-instance and premature-stop failures.}
  \label{fig:vis2}
\end{figure}
Qualitative examples illustrate how EvoMemNav's budgeted coarse-to-fine policy and structured verification improve robustness. As shown in Fig.~\ref{fig:vis2}, we compare EvoMemNav with the baseline across text, image, and object goals. Leveraging VSMGraph retrieval and topological planning, our method avoids the wrong-instance selection and target missing prevalent in baselines with unstructured memory.

%% file: sec/5_conclusion.tex
% \section{Conclusion}
% EvoMemNav is a training-free zero-shot embodied navigation framework built around three components:
% (1) a Visual-Semantic Memory Graph that organizes room–view–object structure while retaining image-grounded view evidence, where view nodes are partitioned into anchors and frontiers for region-level retrieval and multi-view evidence aggregation;
% (2) a budgeted coarse-to-fine policy on the graph, where the coarse Explore stage performs room-aware candidate compression and routes the agent to either a frontier (exploration) or an anchor view (evidence), and the fine Search/Verify stage queries a VLM only over the compact candidate set and performs structured STOP verification to reduce wrong-instance and premature-stop errors;
% (3) Reflection-Driven Continual Memory Adaptation (RDCMA), which writes back subtask outcomes as lightweight, goal-conditioned priors attached to graph elements and updates them across subtasks without updating model parameters, enabling stable self-improvement in the lifelong setting.

% By combining fine-grained image evidence with explicit topological structure and budgeted reasoning, EvoMemNav improves disambiguation, calibrates STOP decisions, and achieves better performance-efficiency trade-offs. Experiments on GOAT-Bench and HM3D across object, text, and image goals show consistent gains in SR/SPL, with fewer premature stops and stronger zero-shot generalization.

\section{Conclusion}
We presented EvoMemNav, a training-free zero-shot navigation framework that (i) maintains a room-view-object VSMGraph with \emph{first-class raw RGB view evidence}, (ii) uses a budgeted coarse-to-fine policy---with coarse = \emph{Explore} and fine = \emph{Search} + \emph{Verify}---for structured multi-view \textsc{Stop} verification, and (iii) performs reflection-driven write-back (RDCMA) to update graph-attached priors without retraining. Experiments on GOAT-Bench and HM3D demonstrate consistent SR/SPL gains across object, text, and image goals, with improved multi-instance disambiguation and fewer premature stops under favorable performance-efficiency trade-offs.

% Limitations and future work. Our study is limited to static indoor scenes in Habitat with accurate depth and pose, and the current implementation still requires relatively expensive VLM queries. Exploring lighter backbones, noisy real-world sensing, and broader benchmarks is a promising direction for future work.